\documentclass[10pt,twocolumn,letterpaper]{article}

\usepackage[pagenumbers]{cvpr} 

\newcommand{\change}[1]{{#1}}

\newcommand{\method}{STORM\xspace}
\newcommand{\dataset}{STORM-Bench\xspace}

\newcommand{\headline}[1]{\noindent\textbf{#1}}

\definecolor{cvprblue}{rgb}{0.21,0.49,0.74}
\usepackage[pagebackref,breaklinks,colorlinks,allcolors=cvprblue]{hyperref}

\def\confName{CVPR}
\def\confYear{2026}

\title{STORM: End-to-End Referring Multi-Object Tracking in Videos}

\author{
Zijia Lu\thanks{Work conducted during an internship at Amazon.}
\quad
Jingru Yi \quad Jue Wang \quad Yuxiao Chen \quad Junwen Chen \quad Xinyu Li \quad Davide Modolo\\
Amazon\\
{\tt\small lu.zij@northeastern.edu \quad \{jyijingr,juewangn,cyuxiao,chenjunw,xxnl,dmodolo\}@amazon.com}
}

\begin{document}
\maketitle

\begin{abstract}
Referring multi-object tracking (RMOT) is a task of associating all the objects in a video that semantically match with given textual queries or referring expressions. 
Existing RMOT approaches decompose object grounding and tracking into separated modules and exhibit limited performance due to the scarcity of training videos, ambiguous annotations, and restricted domains. In this work, we introduce \textbf{\method}, an end-to-end MLLM that jointly performs grounding and tracking within a unified framework, eliminating external detectors and enabling coherent reasoning over appearance, motion, and language. To improve data efficiency, we propose a task-composition learning (TCL) strategy that decomposes RMOT into image grounding and object tracking, allowing STORM to leverage data-rich sub-tasks and learn structured spatial--temporal reasoning. We further construct \textbf{\dataset}, a new RMOT dataset with accurate trajectories and diverse, unambiguous referring expressions generated through a bottom-up annotation pipeline. Extensive experiments show that STORM achieves state-of-the-art performance on image grounding, single-object tracking, and RMOT benchmarks, demonstrating strong generalization and robust spatial--temporal grounding in complex real-world scenarios. \dataset is released at \href{https://github.com/amazon-science/storm-referring-multi-object-grounding}{https://github.com/amazon-science/storm-referring-multi-object-grounding}.
\end{abstract}

\section{Introduction}
\label{sec:intro}
Identifying and localizing objects in videos is essential for various vision applications, including video understanding, human--object interaction analysis, and embodied perception. Precise spatial--temporal localization enables models to reason about how objects move and interact over time. In practice, however, users are typically interested in only a subset of objects relevant to their intent. This motivates the task of \textit{referring object grounding}~\cite{wu2023referring,liu2024grounding,li2024groundinggpt}, where the model detects objects that semantically matched with given referring expressions. Language offers a flexible interface for specifying appearance, attributes, and spatial relations, making referring grounding more practical than category-based detection or generic tracking. 

Early research efforts \cite{chen2023shikra,liu2024grounding,you2023ferret,chen2023minigpt} primarily addressed referring single-object grounding before extending the idea to tracking, where a model localizes and follows one target based on a textual query. While approaches like referring single-object tracking (RSOT)~\cite{sun2024chattracker,wang2024elysium} show strong performance in language-guided tracking, they are inherently limited to a single target and fail to handle multiple interacting objects or relational descriptions. To this end, we explore the more challenging problem of referring multi-object tracking (RMOT) in this work, which seeks to ground and track all objects referenced by textual queries throughout a video. 

\begin{figure*}
    \centering
    \includegraphics[width=\linewidth]{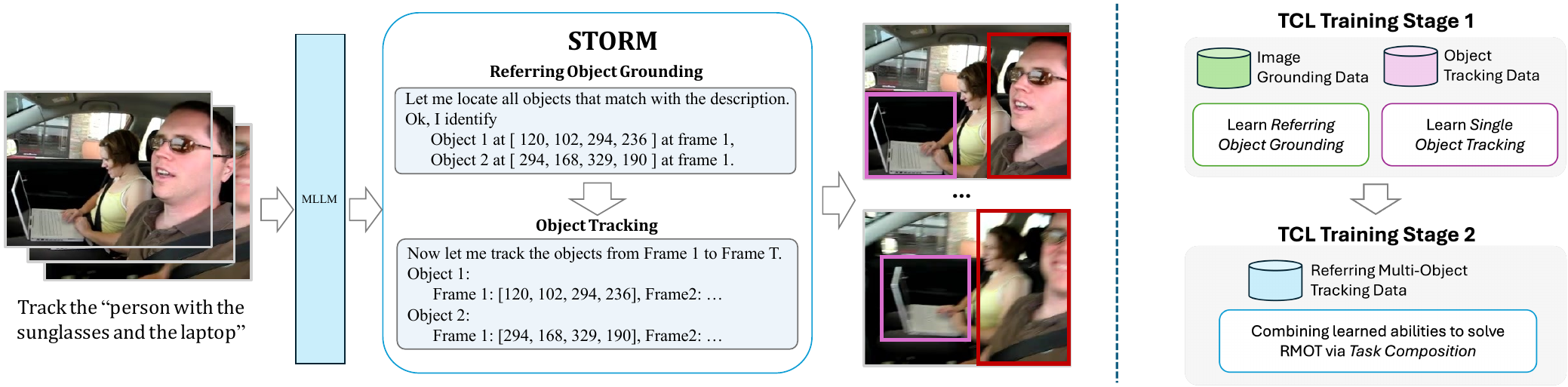}
    \caption{Overview of \method method. The \method adopts a LLaVA-style MLLM architecture~\cite{llava}, enabling video understanding on top of a large language model (LLM). Due to the scarcity of RMOT data, we propose a Task-Composition Learning (TCL) recipe. As illustrated in the figure on the right, TCL employs a two-stage training process. Stage 1 transfers subtask knowledge (i.e., object grounding and single-object tracking, SOT) to the referring object tracking task. In Stage 2, the model is fine-tuned on our proposed \dataset dataset, further enhancing performance on the referring multi-object tracking (RMOT) task with limited videos.}
    \label{fig:enter-label}
    \vspace{-5mm}
\end{figure*}
 
Prior RMOT approaches \cite{zhang2021fairmot,bewley2016simple,wojke2017simple,bergmann2019tracking,zhang2022bytetrack,liang2022rethinking} typically augment a specialized detector with text encoders, but such systems struggle to interpret complex referring expressions and cannot reason about causal or relational dependencies conveyed in language.
Large language models (LLMs), in contrast, excel at understanding nuanced linguistic structures. Most Multi-modal large language models (MLLMs)-based grounding methods~\cite{sun2024chattracker,wang2025reasoningtrack} operate on static images, and its extensions~\cite{chamiti2025refergpt, bai2025qwen2} to videos typically attach an external tracker or detector. However, these methods leverage the LLMs as standalone text encoder and require additional detection module, preventing the model from learning unified spatial-temporal representations. As a result, no existing approach tackle the RMOT in an end-to-end manner.

To fill this gap, we introduce the \textbf{S}patial--\textbf{T}emporal \textbf{O}bject \textbf{R}eferential \textbf{M}odel (\method), the first end-to-end multi-modal large language model designed specifically for referring multi-object tracking (RMOT). STORM unifies grounding and tracking within a single MLLM framework, without relying external detectors or trackers. The model utilizes a ViT-based visual encoder to extract spatial features from individual frames, while the LLM leverages these features together with the text query to capture temporal dynamics and reason about cross-modal correspondence between linguistic descriptions and visual entities. It then directly generates bounding boxes for the text-described objects in a structured plain-text format, allowing the model to fully leverage pretrained language reasoning while avoiding the complexity and computational cost of additional external modules.

As is pointed out in the recent studies~\cite{kaplan2020scaling, hoffmann2022training, amirloo2024understanding}, multi-model large language models (MLLMs) are known to require massive amounts of high-quality training data to achieve strong performance. However, constructing large-scale datasets for referring multi-object tracking is highly impractical. Inspired by perception learning in the LLM pre-training (PT) stage and skill acquisition in the supervised fine-tuning (SFT) stage~\cite{liu2023visual}, we argue that RMOT can likewise be approached by first learning fundamental capabilities from easily accessible data, followed by a small amount of task-specific fine-tuning. To this end, we propose \textit{task-composition learning (TCL)} strategy in STORM, which decompose RMOT as two fundamental tasks: grounding objects according to a textual description and maintaining their identities across time. Specifically, STORM is first trained on large-scale datasets for image grounding and single-object tracking to learn robust cross-modal alignment and temporal consistency. We then fine-tune the model with reasoning-based supervision that guides it to identify all referenced objects in the initial frame and subsequently track them over time. This training paradigm reduces reliance on RMOT-specific annotations and enhances generalization to complex, relational, and ambiguous referring expressions.


Although TCL significantly reduces the need for RMOT-specific data, existing RMOT datasets remain far from satisfactory: they are small in scale, noisy, and lack sufficient diversity. To address these limitations, we construct \dataset, a new dataset featuring accurate multi-object trajectories and diverse, unambiguous referring expressions. \dataset is created through a bottom-up annotation pipeline that first generates and verifies object-level descriptions, and then composes multi-object expressions via controlled language reasoning. This process ensures that the final expressions capture the object attributes, spatial relationships, and temporal interactions essential for referring multi-object tracking.

%

In summary, our main contributions in this work are as follows:
\begin{itemize}
    \item We propose \method, the first end-to-end MLLM framework for referring multi-object tracking that eliminates the need for external grounding modules.
    \item We introduce a task composition learning (TCL) strategy—a simple yet effective framework that substantially reduces the reliance on RMOT-specific data. TCL enables an MLLM to acquire RMOT capabilities by first learning from widely available subtask datasets and then applying a small amount of RMOT-focused fine-tuning.
    \item We introduce \dataset, a high-quality and challenging dataset for referring multi-object grounding task, which includes 0.2M diverse referring expressions and 73.7K tracked objects.
    \item We show STORM achieves the state-of-the-art performance on various benchmarks of object grounding, single object tracking, and referring multi-object tracking.
\end{itemize}

\section{Related Works}
\label{sec:formatting}
\headline{Object Grounding.}
Object grounding is a task that precisely localizes objects referred in a given natural language from images or videos. Object detection-based methods (such as MDETR \cite{kamath2104modulated}, GLIP \cite{li2022grounded}, Grounding DINO \cite{liu2024grounding}) align object and text features in embedding space and subsequently retrieve object-text pairs. To support timestamp detection, Video Temporal Grounding (VTG) \cite{Lu:ICCV21,lin2023univtg,Lu:CVPR22,caba2015activitynet,lei2021detecting,Lu:CVPR24,Lu:CVPR25,Lu:ICCV25}) are proposed to further extend grounding task for videos and support substream tasks such as time retrieval, video summarization, highlight detection. In recent years, LLM-based visual-language models have integrated spatial-temporal grounding (STG) task and serve for zero-shot grounding with complex languages
\cite{wang2024elysium,guo2025vtg,huang2024vtimellm,wu2025number,uzkent2023dynamicinference}. Object grounding with precise coordinates are further introduced in MLLM models \cite{wang2024qwen2,bai2025qwen2,chen2023minigpt,wang2024elysium,chen2023shikra,peng2023kosmos,zhang2024gpt4roi}. Open-vocabulary approaches extend grounding to unseen categories by exploiting VLM localizability and semantics \cite{bao2025openvocabaction}.

\headline{Referring Object Tracking.} Object tracking associates objects across video frames without specific user guidance. Referring Object Tracking merges object grounding and object tracking and only tracks objects specified in input textual queries \cite{chamiti2025refergpt,wu2023referring}. In traditional object tracking, similarity-learning is commonly adopted in single object tracking (SOT) task \cite{li2018high,bertinetto2016fully,li2019siamrpn++}. Multi-object tracking  (MOT) task generally employs tracking-by-detection paradigm and utilizes additional tracker head to associate object bounding boxes \cite{zhang2021fairmot,bewley2016simple,wojke2017simple,bergmann2019tracking,zhang2022bytetrack,liang2022rethinking,lu2024pathconsistency}. To mitigate the limited object categories and further support referring object tracking, researchers have utilized visual-language models \cite{feng2021siamese,li2023ovtrack,zhou2023joint,yu2023generalizing,li2017tracking} for open-set object association and classification. LLM-based referring object tracking are explored in recent works \cite{sun2024chattracker,wang2025reasoningtrack}. Scalable MLLM-based trackers further build on advances in video representation learning, including motion-aware contrastive pretraining \cite{xiao2022maclr}, text-guided masked autoencoders \cite{fan2024textguidedvideomae}, online temporal action modeling \cite{chen2022gatehub}, and efficient long-sequence processing \cite{wang2023selectiveS5,lee2024videotokenmerging,chen2026compactvideo}. In particular, ReasoningTrack \cite{wang2025reasoningtrack} incorporated a tracking head in Qwen2.5-VL \cite{bai2025qwen2} and iteratively update single object tracklets. ReferGPT \cite{chamiti2025refergpt} proposed a matching module on top of MLLM models to support zero-shot multi-object tracking for cityscape driving scenes. ChatTracker \cite{sun2024chattracker} enhances tracking ability with improved data annotations. Elysium \cite{wang2024elysium} constructs a ElysiumTrack-1M dataset to support SOT and Referring SOT (RSOT) tasks on top of MLLM models. Referring MOT (RMOT) \cite{wu2023referring,chamiti2025refergpt,lin2024echotrack,zhang2024bootstrapping,he2024visual,du2024ikun} associates all objects semantically matched with referring expressions.

\headline{Referring Object Tracking Datasets.} The existing SOT \cite{wu2013online,real2017youtube} or MOT \cite{geiger2012we,milan2016mot16} datasets generally lack referring expressions and have limited categories. Moreover, the small quantity of videos make them inadequate for large-language model finetuning. To support referring understanding, several object grounding datasets are proposed for image-based (Flickr30k \cite{young-etal-2014-image}, RefCOCO~\cite{kazemzadeh2014referitgame}, RefCOCO+~\cite{kazemzadeh2014referitgame}, and RefCOCOg~\cite{mao2016generation}) and video-based  (OTB99 \cite{li2017tracking}, Cityscapes-Ref \cite{vasudevan2018object}, Talk2Car \cite{deruyttere2019talk2car}, Refer-Youtube-VOS \cite{seo2020urvos}) referring tasks.  More recently, RSOT~\cite{fan2019lasot} and RMOT~\cite{li2025lamot} datasets incorporate textual annotations but suffer from annotation quality issues. For instance, LaSOT~\cite{fan2019lasot,wang2024elysium} provides text descriptions for single-object tracking, yet videos may contain multiple instances of the same category, and the referring expression fails to specify which instance should be tracked. LaMOT~\cite{li2025lamot} represents one of the earliest efforts to build MOT datasets with referring expressions but still exhibits low annotation diversity and incomplete object coverage. Elysium~\cite{wang2024elysium} proposes a larger-scale referring single-object tracking dataset based on WebVid videos~\cite{bain2021frozen} while the dataset has bunch of misaligned or erroneous bounding boxes that do not accurately match the referring expression. Existing RMOT works are generally focused on cityscape scenes (Ref-KITTI \cite{wu2023referring,zhang2024bootstrapping}). We propose a new dataset \dataset which corrects the ambiguity in existing RMOT datasets and extends RMOT to various domains.



\section{Referring Multi-Object Tracking}
\label{sec:method}

In this section, we first introduce the problem definition. Next, we present the design of our Spatial-Temporal Object Referring Model (\method). Finally, we describe the construction of our collected dataset (\dataset) in detail.

\subsection{Problem Definition}
Referring multi-object tracking (RMOT) aims to track all objects throughout a video that semantically align with specified textual queries or referring expressions. Given a video $\mathcal{V} = \{I_t,  t\in \{1, \dots, T\}\}$ consisting of $T$ frames and a referring expression $\mathcal{R}$ describing one or more objects of interest, the model extracts the spatial locations $\mathcal{B} = \{B_t^k, t\in \{1, \dots, T\}, k \in \{1, \dots, K\}\}$ of all corresponding objects $K$ across video frames $T$.

%

\subsection{\method}
Unlike existing works \cite{wu2023referring,chamiti2025refergpt,zhang2024bootstrapping,rasheed2024glamm,bai2024one,lin2025glus} that decompose grounding and tracking into separate modules, we unify RMOT within a single multi-modal large language model (MLLM) framework, leveraging the strong reasoning capabilities of LLMs and maintain semantic consistency of objects across temporal frames. We name our proposed model as \method. In addition, due to scarcity of RMOT datasets, we design a task-composition learning (TCL) method that allows model to leverage knowledge learned from sub-tasks such as image grounding and SOT, and thereby generalize to RMOT task.


\noindent\textbf{Architecture.} \method's model architecture follows the common LLaVA-style MLLM design~\cite{llava}. In particular, \method extracts the frame-based visual tokens $\mathcal{V}$ through a ViT-based vision encoder. A two-layer MLP projector is followed to map the visual tokens into text space. The constructed visual tokens and referring text query tokens are sent to the a LLaMA-based LLM~\cite{llava} which auto-regressively generates the RMOT outputs $\mathcal{B}$. This formulation enables us to train the model as a next-token prediction task using cross-entropy loss.   

\noindent\textbf{Prompt and RMOT Output Format.} We adopt a consistent prompting format for $\mathcal{R}$: \textit{``\textless video\textgreater~Please locate all objects in the video based on this expression: [referring expression].''}. The RMOT output response is formated as: \textit{``Object 1: Frame 1: [x1, y1, x2, y2], Frame 2: [x1, y1, x2, y2], ...; 
Object 2: ...''}, where $(x_1, y_1, x_2, y_2)$ denotes the absolute coordinates of a bounding box. When objects are temporarily absent from a frame due to occlusion or camera movement, \method outputs an empty bounding box to indicate that the object is unobserved in that frame.
\change{For long videos, we split the input into shorter clips and stitch the resulting tracklets by using the predicted boxes from the previous clip as prompts for the next one.}


\noindent\textbf{Task-Composition Learning (TCL).} Training an MLLM for a new task typically requires large-scale annotated datasets. However, collecting large-scale RMOT training data is prohibitively expensive, as it demands frame-level object annotations paired with natural language descriptions that precisely reference specific objects over time. To address this challenge, we adopt a task-composition strategy that decomposes the complex RMOT task into simpler subtasks with abundant data, allowing the model to be trained on large-scale datasets for each component task. Specifically, 
we first train the model on extensive object grounding and object tracking datasets to establish strong visual–linguistic and temporal tracking capabilities. We then progressively extend the model’s competence to RSOT and ultimately to RMOT using smaller, task-specific datasets. This staged training pipeline enables effective RMOT performance while significantly reducing the need for large-scale RMOT annotations. The detailed TCL training recipe is provided below. In particular, our training process consists of two stages:

\textit{Stage1: Pretraining on image grounding and object tracking.}  
We train \method\ on large-scale referring image grounding datasets, enabling it to associate textual expressions with spatial visual regions and predict object bounding boxes in single frames. This stage equips the model with strong cross-modal alignment between appearance cues and linguistic descriptions. In addition, we incorporate large-scale single-object tracking (SOT) datasets to help the model learn temporal consistency and motion representations across frames. Finally, we incorporate referring single-object tracking (RSOT) datasets, which help the model integrate its learned visual–linguistic alignment with temporal tracking capabilities, effectively teaching the LLM to perform language-guided single-object tracking.


\textit{Stage2: Finetuning with \dataset.}  
Finally, we finetune the model using our curated \dataset dataset.  
To encourage explicit reasoning, we employ a \textit{Chain-of-Thought (CoT)} training strategy.  The model is guided to generate intermediate reasoning traces, first grounding the described objects in the initial frame and then tracking them sequentially throughout the video.  This thinking process allows the model to learn the structured composition of the two sub-tasks, leading to improved performance and interpretability. After the reasoning phase, the model outputs object locations following the structured format described above. 

This multi-stage training scheme in TCL allows the model to effectively transfer knowledge from abundant image grounding and tracking datasets, while requiring only a modest amount of referring multi-object data for adaptation. Empirically, we find that TCL substantially improves generalization to complex referring expressions and enhances robustness to ambiguous or previously unseen object descriptions.

\begin{figure}
    \centering
    \includegraphics[width=\linewidth]{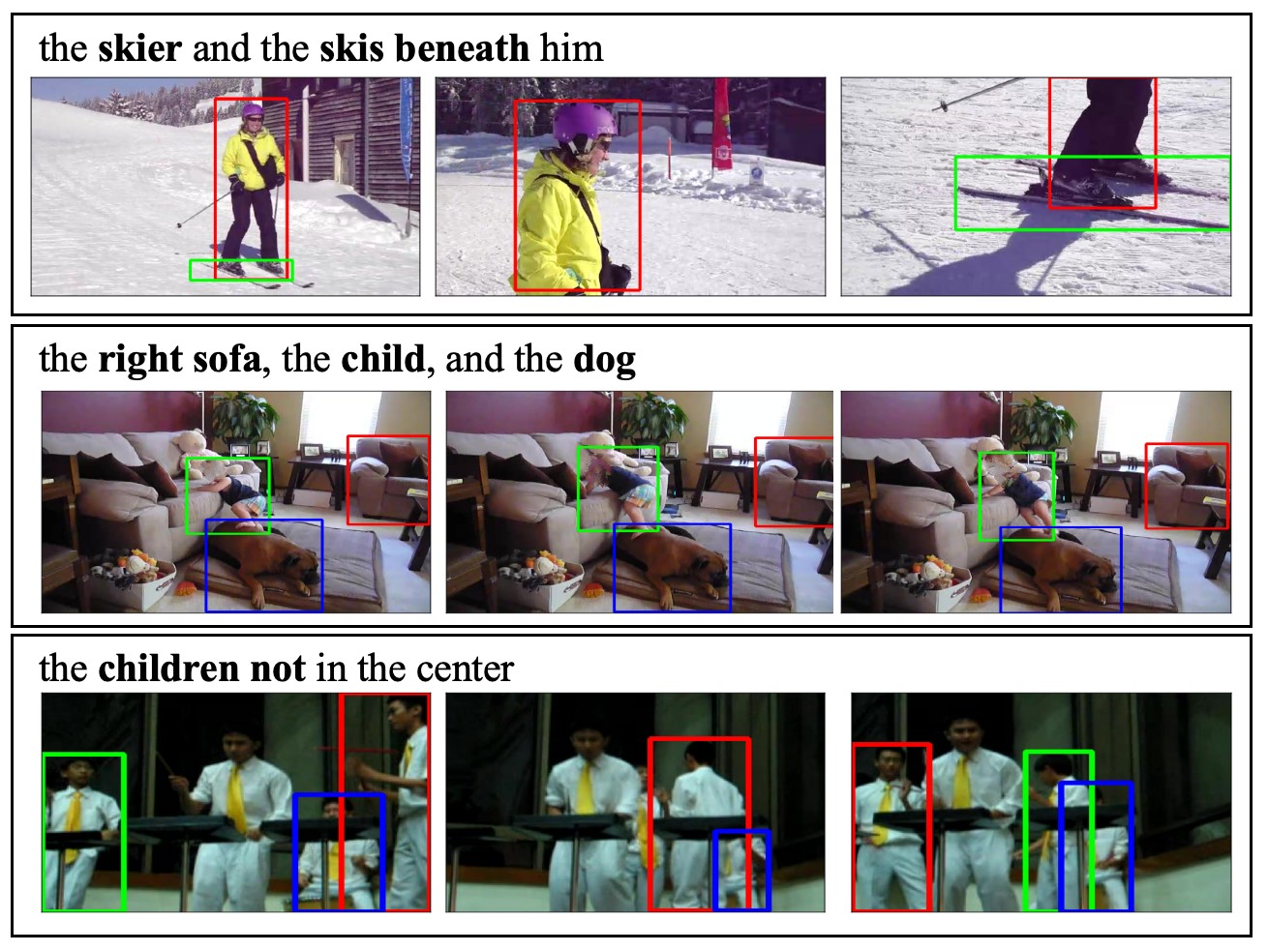}
    \caption{Visualization of videos from our \dataset\ dataset. Bounding boxes in different colors represent different tracklets. The dataset includes a wide range of challenging scenarios: top—significant scale changes for the same target; middle—multi-object tracking with clear prompts under weaker conditions; and bottom—multi-target tracking of similar object types but different instances under stronger, more specific conditions.}
    \label{fig:dataset-vis}
    \vspace{-5mm}
\end{figure}
\label{sec:dataset}

\section{\dataset}
\label{sec:method}

\subsection{Dataset Construction}
Training and evaluating the proposed \method model requires a dataset with high-quality referring expressions and accurate object bounding boxes across video frames. While our task-composition learning (TCL) strategy reduces the need for large-scale RMOT training data, using existing RMOT datasets \cite{li2025lamot} may yield limited performance due to ambiguous and inaccurate descriptions of the target objects \cite{sun2024chattracker}. In this work, we introduce a new dataset \dataset, featuring carefully designed referring expressions to support both the training and evaluation of referring multi-object tracking methods.








\vspace{3pt}
\noindent\textbf{Bottom-up annotation pipeline.}
To support tracking tasks in MLLM, existing work Elysium \cite{wang2024elysium} proposed a \textit{top-down} pipeline for million-scale RSOT data annotations:
First, it obtains the caption of a video and generates bounding boxes for the nouns in the caption by Grounding-DINO. 
Second, the generated bbox-object pairs are used to create tracklets of each object across frames by a tracking model e.g. Mixformer. The top-down pipeline is highly sensitive to the performance of the expert detector and the tracker. 
In contrast, we propose a \textit{bottom-up} data annotation pipeline. We leverage the asymmetry in task difficulty—while locating an object given a description is challenging, generating captions for a given object is a relatively easy task. The bottom-up pipeline first localizes objects and then generates corresponding referring expression for each object. Unlike Elysium~\cite{wang2024elysium} that simply uses non-semantic confidence scores for box and tracklets filtering, we employ an LLM-based verification process that leverages the reasoning capabilities of large language models to identify and filter incorrect annotations. The \dataset dataset is built upon the Vidor dataset~\cite{shang2019annotating,thomee2016yfcc100m}, which provides ground-truth bounding boxes for diverse daily objects, beyond the limited pedestrian–vehicle focus of typical MOT datasets~\cite{wu2023referring,zhang2024bootstrapping}. The bottom-up pipeline consists of two main stages: object-level caption generation and verification, and multi-object referring expression synthesis and validation.

\begin{table}[]
    \centering
    \resizebox{\linewidth}{!}{%
    \begin{tabular}{c|c|c|c|c|c}
    \toprule
            &  \# Expressions & \# Words & \# Length & MO & CE\\ \midrule
    LaSOT \cite{fan2019lasot}  & 1.4K & 9.8K & 7.0 & $\times$ & $\times$ \\ 
    Elysium \cite{wang2024elysium}& 1.3M & 2.7M & 2.1 & $\times$ & $\times$ \\
    Refer-KITTI-V2 \cite{refer-kitti-v2} & 9.8K & 63.9K & 6.6 & $\checkmark$ & $\times$ \\
    LaMOT \cite{li2025lamot}  & 7.5K & 28K  & 3.8 & $\checkmark$  & $\times$ \\ \midrule
    Ours    & 0.2M & 2.4M & 12.2 & $\checkmark$ & $\checkmark$ \\ \bottomrule
    \end{tabular}
    }
    \caption{\dataset dataset statistics and comparison with other referring tracking datasets. MO: Multi-Objects; CE: Complex-expression.}
    \vspace{-5mm}
    \label{tab:dataset}
\end{table}

\vspace{3pt}
\noindent\textit{Stage1: Object-level caption generation and verification.}
Stage1 aims to generate detailed captions for each object in a video. To get the object-level captions, we feed the MLLM~\cite{bai2025qwen2} model with the video frames and the bounding boxes corresponding to the target object. 
In addition, we adopt visual prompting techniques~\cite{wu2025number,wu2024visual} to enhance model's attention to the specific object, in which we draw
a red bounding box around the target object as a visual marker. 
We find that visual prompts substantially improve caption quality; however, occasional inaccuracies still arise due to inherent hallucinations in MLLMs. 

To address this, we introduce a verification process to detect and filter incorrect captions. In particular, we ask another MLLM to predict whether the generated caption uniquely matches the intended object. 
During verification, we vary the visual inputs as: (1) the original video with red bounding boxes, (2) videos with Gaussian-blurred backgrounds outside the boxes, and (3) cropped object patches. The three types of input are independently evaluated and only the captions verified across all settings are retained. Notably, we found that blurred or cropped inputs are ineffective during caption generation, as removing contextual cues leads to inaccurate descriptions.

\vspace{3pt}
\noindent\textit{Stage2: Multi-object referring expression synthesis and validation.}
After obtaining verified captions for individual objects, we proceed to generate referring expressions that jointly describe multiple objects in the same video. These expressions can take two main forms: (1) a single phrase referring to a group of objects sharing a common attribute (\textit{e.g.}, ``the instruments on the shelf''), or (2) a conjunction of phrases describing distinct objects (\textit{e.g.}, ``the girl and the cat running behind her''). To generate such expressions, we employ a textual LLM with reasoning capabilities. 
We list all of objects in the video with indices and construct the prompt as the object indice and the corresponding caption pairs. 
Along with the indices of a randomly sampled subset, we ask LLM to identify 1–5 shared and distinctive attributes and to compose concise referring expressions. If no such attributes are found, the model is allowed to return an empty output. For conjunction-type expressions, the model produces individual object phrases and then combines them into a single sentence.

Similar to Stage I, we apply a verification step to ensure the multi-object referring expressions are clear and correct. We take the video, bounding box coordinates, and the generated expression into an MLLM to check for semantic consistency and filter out the mismatched object/expression pairs. In this stage, we omit visual prompts such as bounding box overlays, as they tend to clutter the scene and obscure objects when multiple targets are present.

In summary, our two-stage bottom-up annotation pipeline enables the generation of high-quality referring expressions grounded in multiple object locations, forming a large-scale dataset suitable for training and evaluating the \method\ model.


\subsection{Dataset Statistics}
Leveraging our annotation pipeline, we curated a new dataset \dataset with 15093 training videos and 714 evaluation videos, with 0.2M diverse referring expressions and 73.7K tracked objects. We visualize example videos in Figure \ref{fig:dataset-vis}. It can be observed that the dataset includes diverse patterns of referring expressions, such as referring objects by spatial-temporal relations or certain unique attributes, or referring to a list of different objects.

We also compare our dataset with recent popular object tracking datasets (LaSOT~\cite{fan2019lasot}, Elysium~\cite{wang2024elysium}, LaMOT~\cite{li2025lamot}, and Refer-KITTI-V2~\cite{refer-kitti-v2}) in Table~\ref{tab:dataset}. LaSOT and Elysium are designed for referring single-object tracking with simple expressions and are therefore not suitable for our task.
\change{Refer-KITTI-V2 is one of the first referring multi-object tracking datasets. However, it is limited to two object classes, traffic scenes and simple expressions.}
LaMOT is another referring multi-object tracking dataset with diverse scenes. Yet, it is relatively small in scale and many videos use simple object categories as the referring expressions.
In contrast, \dataset contains 15.7K videos and 251K images, averages 3 instances per expression, and spans 80 classes with an average expression length of 12.2.
Together, these statistics highlight the broader domain coverage and richer relational language in \dataset.

\label{sec:method}


\section{Experimental Results}
\label{sec:experiments}
\subsection{Experimental Settings}
\headline{Benchmarks.}
We adopt a progressive evaluation protocol: beginning with image object grounding, followed by referring single-object tracking, and ultimately referring multi-object tracking, to provide a comprehensive assessment of our method.
For the \textit{image object grounding} task, we use the widely adopted RefCOCO~\cite{kazemzadeh2014referitgame}, RefCOCO+~\cite{kazemzadeh2014referitgame}, and RefCOCOg~\cite{mao2016generation} benchmarks.
RefCOCO and RefCOCO+ are built on MS-COCO and provide region-level referring expressions for objects in static images, RefCOCO emphasizing spatial reasoning and RefCOCO+ focusing on appearance-based descriptions.
RefCOCOg contains longer, more descriptive expressions, making it suitable for assessing language understanding in complex scenes. Following \cite{wang2024elysium}, we report results on the \textit{val, testA, and testB} splits.
We evaluate the referring single-object tracking (RSOT) task on Elysium~\cite{wang2024elysium} test dataset. Elysium~\cite{wang2024elysium} contains videos sourced from WebVid-10M~\cite{bain2021frozen}, the dataset provides 500 test videos with concise referring expressions and tracking annotations. 
Finally, we evaluate the referring multi-object tracking (RMOT) task on the proposed \dataset dataset. 
Following standard multi-object tracking (MOT) protocols, we report RMOT performances with HOTA~\cite{luiten10higher}, identity-based metrics~\cite{idf1}, and CLEAR metrics~\cite{clear-metric} (MOTA and IDsw).

\headline{Implementation Details.}
\change{Our \method\ model is built on an 8B base-model~\cite{liu2023visual}, with a NaViT~\cite{dehghani2023patch} vision encoder.  
We choose NaViT because it naturally supports arbitrary resolutions and aspect ratios without aggressive resizing, and because it is more token-efficient than common multi-crop ViT alternatives for grounding-heavy video inputs. }
We use publicly released weights from Hugging Face and finetune the model for image grounding, RSOT, and RMOT following our task-composition training.

\subsection{Main Results}
\method builds on MLLM model, leveraging its strong text comprehension and spatial understanding capabilities. In addition, our proposed TCL strategy alleviates the need for large-scale RMOT data and further enhances the RSOT and RMOT performance of \method. In this section, we present a thorough evaluation of \method by decomposing its performance across individual subtasks. Unless otherwise noted, we report the best results obtained from models trained on image grounding, SOT, RSOT, and RMOT data.

\headline{Image Grounding Results.}
RSOT and RMOT fundamentally depend on strong spatial grounding, so we first examine whether \method\ retains this capability on images, despite being primarily designed for video grounding and tracking.
Table~\ref{tab:image-grounding} presents comparisons with existing MLLMs on RefCOCO, RefCOCO+, and RefCOCOg.
Across all datasets and evaluation settings, \method achieves state-of-the-art performance, demonstrating robust generalization across grounding tasks and visual modalities.
Notably, our model is trained on no additional grounding data beyond RefCOCO, the same dataset used by all baselines, suggesting that the image grounding performance also benefits from the additional RSOT and RMOT data used in our training framework.

\begin{table}[t]
    \footnotesize
    \centering
    \begin{tabular}{l|c|c|c}
    \toprule
    & \multicolumn{1}{c|}{RefCOCO} & \multicolumn{1}{c|}{RefCOCO+} & \multicolumn{1}{c}{RefCOCOg} \\
    & val/testA/testB & val/testA/testB & val/test \\ \midrule
 Shik-7B\cite{chen2023shikra}   & 87.0/90.6/80.2 & 81.6/87.4/72.1 & 82.3/82.2 \\
 Shik-13B\cite{chen2023shikra}  & 87.8/91.1/81.8 & \textbf{82.9}/87.8/74.4 & 82.6/83.2 \\
 M-GPT2\cite{chen2023minigpt}  & 88.7/91.7/\textbf{85.3} & 80.0/85.1/\textbf{74.5} & \textbf{84.4}/84.7 \\
 Ferret\cite{you2023ferret}        & 87.5/91.4/82.5 & 80.8/87.4/73.1 & 83.9/84.8 \\
 G-GPT\cite{li2024groundinggpt} & 88.0/91.6/82.5 & 81.6/ 87.2/73.2 & 81.7/82.0 \\ \midrule
 Ours & \textbf{89.1}/\textbf{92.7}/83.8 & 81.6/\textbf{88.6}/73.5 & 84.0/\textbf{85.1} \\ 
    \bottomrule         
    \end{tabular}
    \caption{Referring image object grounding performance on RefCOCO, RefCOCO+, and RefCOCOg.}
    \label{tab:image-grounding}
    \vspace{-2mm}
\end{table}

\begin{table}[b]
\centering
\footnotesize
\begin{tabular}{l|ccc|ccc}
\toprule
 & \multicolumn{3}{c|}{\begin{tabular}[c]{@{}c@{}}Elysium\\ (RSOT)\end{tabular}} & \multicolumn{3}{c}{\begin{tabular}[c]{@{}c@{}}Elysium\\ (SOT)\end{tabular}} \\ \midrule
 & AUC & P &  P\textsubscript{norm} & AUC & P &  P\textsubscript{norm} \\ \midrule
Elys~\cite{wang2024elysium} & 83.3 & 89.1 & 90.0 & 88.7 & 94.6 & 93.8  \\
Ours & \textbf{84.1} & \textbf{89.7} & \textbf{93.2} & \textbf{89.8} & 
\textbf{96.4} & \textbf{97.8}  \\ 
\midrule
Elys*  & \textbf{87.5} & 94.5 & 93.7 & 88.7 & 94.6 & 93.8  \\
Ours* & 87.4 & \textbf{95.3} & \textbf{97.2} & \textbf{89.8} & \textbf{96.4} & \textbf{97.8}  \\ 
\bottomrule
\end{tabular}
\caption{Single-object tracking (SOT) and referring single-object tracking (RSOT) results on Elysium. * denotes the experiments with our prompts (longer and more comprehensive).}
\label{tab:SOT}
\end{table}


\begin{table*}[t!]
\centering
\resizebox{0.93\linewidth}{!}{%
\begin{tabular}{l|cccc|ccc|cc}
\toprule
{} & {\textbf{HOTA}$\uparrow$} & {\textbf{DetA}$\uparrow$} & {\textbf{AssA}$\uparrow$} & {\textbf{LocA}$\uparrow$} & {\textbf{IDF1}$\uparrow$} & \textbf{IDP}$\uparrow$ & \textbf{IDR}$\uparrow$ & \textbf{MOTA}$\uparrow$ & \textbf{IDsw}$\downarrow$ \\ \midrule
 {Grounding DINO~\cite{li2022grounded}$\dagger$}             &  31.7 & 17.8 & 56.5 & 88.1 & 27.3 & 68.5 & 17.0 & 15.4 & 2953 \\
 {Qwen2.5-VL~\cite{bai2025qwen2} $\dagger$}                 &  37.9 & 23.2 & 62.6 & 76.6 & 38.6 & 77.9 & 25.5 & 22.4 & 2767 \\
 {VisionLLMv2~\cite{wu2024visionllmv2}$\dagger$}                &  45.3 & 35.5 & 58.9 & 75.3 & 44.6 & 53.9 & 38.1 & 18.7 & 12097 \\
{LaMOT}~\cite{li2025lamot}                               &  46.7 & 39.0 & 56.1 & 89.2 & 48.8 & 71.5 & 37.1 & 37.6 & 7972 \\ \midrule
 {Ours*}  &  34.8 & 21.4 & 56.8 & 63.7 &  36.4 & 54.2 & 27.3 & 15.9 & 2832 \\ 
  {Ours}  &  \textbf{66.7} & \textbf{55.3} & \textbf{82.6} & \textbf{81.4} & \textbf{78.3} & \textbf{78.4} & \textbf{78.1} & \textbf{57.1} & \textbf{183} \\
 \bottomrule
\end{tabular}
}
\caption{Referring multi-object tracking (RMOT) performance on \dataset. $\dagger$ indicates that OCSORT is used as the tracker. *denotes the model trained on \dataset only.}
\label{tab:MOT}
\vspace{-5mm}
\end{table*}

\headline{Referring Single-Object Tracking.}
Following prior work~\cite{wang2024elysium}, we evaluate RSOT under two settings: \textit{referring tracking} setting, where the model must track an object given a natural-language referring expression, and \textit{standard tracking} setting, where the object is specified by its initial-frame bounding box.
We observed that referring tracking performance is sensitive to the quality of the referring expressions. In Elysium~\cite{wang2024elysium}, most expressions are short noun phrases. To further test the model, we use an LLM to rewrite these phrases into longer and more descriptive expressions, applying the same process to both \method\ and Elysium to further test the model's the performance on long and complex prompts.
The results (Table~\ref{tab:SOT}) show the \method\ performs similarly comparing to the Elysium with original short prompts. But with our enriched prompts, the \method\  performs strongly in both evaluation settings, improving AUC by 0.8\% and 2.1\%, respectively. This demonstrates that our task composition strategy enables effective knowledge transfer from object grounding to tracking. 

\headline{Referring Multi-Object Tracking.}
We finally evaluate \method\ on the \dataset\ dataset, the first benchmark specifically designed for RMOT, featuring complex referring expressions and challenging multi-object scenarios that stress-test a model’s ability to understand user queries.
Because no existing model supports end-to-end RMOT, we selected strong baselines and adapted them to operate in the RMOT setting. Image-grounding capable models such as Grounding DINO~\cite{liu2024grounding}, Qwen2.5-VL~\cite{bai2025qwen2}, and VisionLLMv2~\cite{wu2024visionllmv2} can produce bounding boxes per frame given a prompt, but cannot track objects over time. For these models, we generate frame-level detections and apply the linking algorithm~\cite{ocsort} to form trajectories for fair comparison.
LaMOT~\cite{li2025lamot} is a specialist video model that combines an image detector with a temporal tracking module. For a fair comparison, we use its official implementation and pretrained weights, and additionally fine-tune it on the \dataset training split. 

As shown in Table~\ref{tab:MOT}, Grounding DINO performs poorly in the RMOT setting, reflecting its limitations as an image-only model when handling complex queries. 
Qwen2.5-VL and VisionLLMv2 are largely image-based MLLMs, and therefore cannot effectively leverage temporal information to produce consistent object trajectories across frames. Compared to QwenVL, VisionLLMv2 additionally incorporates an object detector; however, its LLM only provides object embeddings to the detector and does not directly participate in localization. As a result, it can misinterpret expressions and occasionally detect false-positive objects. LaMOT, being a video model, exhibits stronger temporal consistency, but it lacks an LLM component and thus struggles to interpret complex referring expressions.

In contrast, \method\ delivers a substantial leap in performance, surpassing all prior methods by +21.2 HOTA and +31.2 IDF1, demonstrating superior spatial-temporal understanding and RMOT capability. Its end-to-end design allows seamless integration of video context with complex natural-language queries. 
Moreover, comparing the results of \textit{Ours*} and \textit{Ours} shows that incorporating TCL improves HOTA by +31.9 and IDF1 by +41.9, validating that TCL effectively reduces the amount of task-specific data needed to learn new tasks. 
Overall, despite the difficulty of large-scale RMOT annotation, our task composition strategy and curated dataset contribute to \method’s robustness and strong generalization in this challenging setting.

\subsection{Ablation Study}

\headline{Task-Composition Learning.}
We conduct ablations to compare RSOT and RMOT performance with and without our proposed task-composition learning strategy, as shown in Table~\ref{tab:composition}. Row 1 establishes the baseline using image-grounding data combined with a small amount of RSOT data.
Rows 2 and 3 demonstrate that our task composition method effectively transfers knowledge from image grounding and SOT to RSOT. Notably, the task-composition–based model achieves performance comparable to RSOT fine-tuning that uses 10× more RSOT data, indicating that our approach is both budget-efficient (no need large scale RSOT annotation) and effective.
A similar trend appears in rows 2 and 4: the task composition strategy also benefits RMOT, and the remaining gap can be closed with only a small amount of RMOT data.

\begin{table}[t]
    \centering
    \footnotesize
    \resizebox{\linewidth}{!}{%
    \begin{tabular}{c|cccc|ccc}
    \toprule
    & \multicolumn{4}{c|}{Training Data Scale} & \multicolumn{3}{c}{Test Task Performance} \\
    ROW & IG & SOT & RSOT & RMOT & SOT &RSOT & RMOT   \\ \midrule
    1 & 1M & 0M & 0.1M &0K& 83.4 &77.4 & 36.9 \\ 
    2 & 1M & 0M & 1M &0K&  89.8 &84.1 & 43.3 \\ \midrule 
    3 & 1M & 0.9M & 0.1M &0K& 91.3 & 83.0 &  41.8 \\ 
    4 & 1M & 0M & 1M &15K &  91.3 & 83.0 & 66.7 \\ 
    \bottomrule
    \end{tabular}}
    \caption{Ablation on task-composition training using image grounding (IG), single-object tracking (SOT), and referring single-object tracking (RSOT) data.``M'' and ``K'' denote million and thousand, respectively.}
    \label{tab:composition}
    \vspace{-3mm}
\end{table}



\begin{figure}
    \centering
    \includegraphics[width=0.85\linewidth]{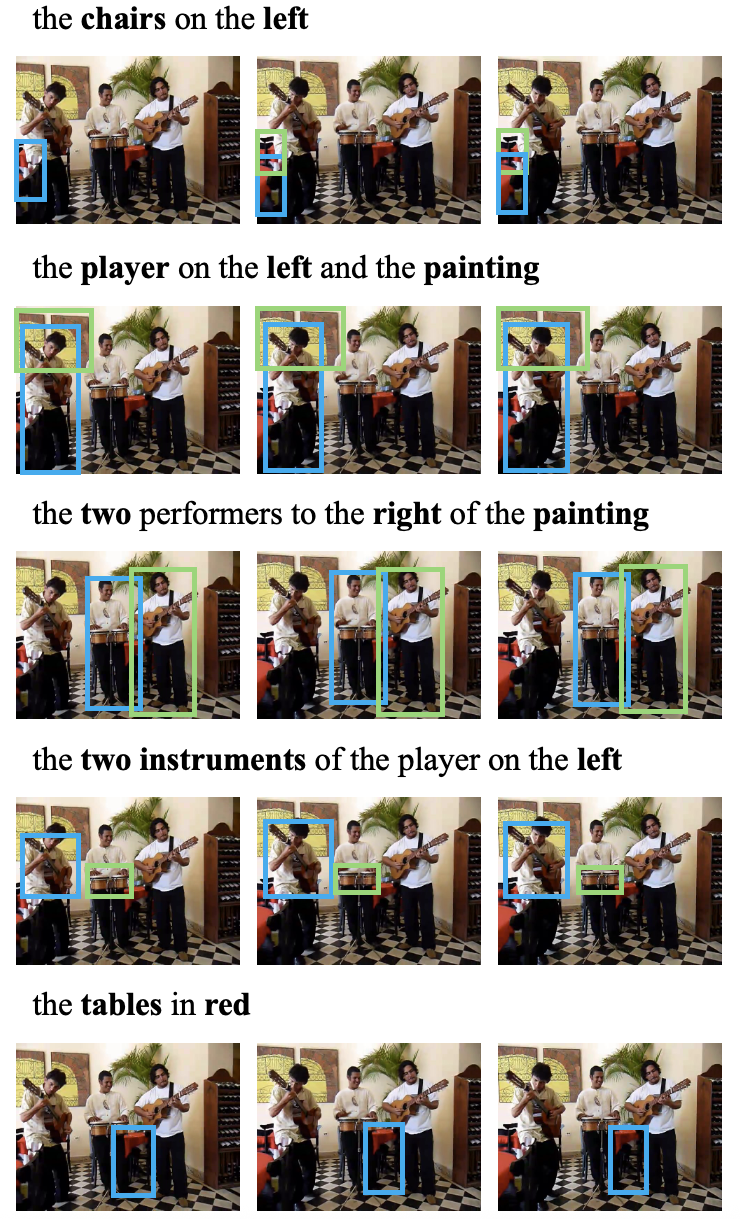}
    \caption{Visualization of complex referring expressions involving attributes and relations (e.g., spatial relations between objects). Our model maintains consistent identities and precise localization across challenging scenarios.}
    \label{fig:vis3}
    \vspace{-2mm}
\end{figure}

\headline{RMOT Scale.}
Finally, we examine how scaling RMOT data impacts performance, noting that RMOT annotations are extremely costly and difficult to obtain. As shown in Table~\ref{tab:rmot-scale}, performance begins to plateau at around 12K samples. Motivated by this trend, we cap our \dataset dataset at 15K samples.

\begin{table}[]
    \centering
    \resizebox{0.78\linewidth}{!}{%
    \begin{tabular}{c|c|c|c|c}
    \toprule
       RMOT Training Data & 5K & 10K & 12K  & 15K  \\ \midrule
       RMOT Performance (HOTA)  &  26.4 &  53.4  &  63.3 & 66.7 \\
       \bottomrule
    \end{tabular}}
    \caption{\small Effect of the scale of RMOT training data in TCL.}
    \label{tab:rmot-scale}
    \vspace{-5mm}
\end{table}

\subsection{Qualitative Results}
First, we demonstrate the generalization ability of \method\ by visualizing tracklets produced from different prompts on the same video (Figure~\ref{fig:vis3}). In this example, we begin with a simple single-object prompt containing basic spatial cues, and then progressively enrich the prompts with additional spatial and relational conditions to track objects of varying locations and scales. \method\ successfully interprets both simple and complex prompts and generates high-quality, consistent tracklets.

We then visualize and compare \method\ with Qwen2.5-VL and VisionLLMv2 on their tracking results to better understand the differences between approaches (Figure~\ref{fig:vis1}).
Qwen2.5-VL~\cite{bai2025qwen2} cannot fully understand the RMOT prompts, and makes the mistakes in the initial grounding generating the inconsistent bounding boxes, resulting in wrong tacklets.
VisionLLMv2~\cite{wu2024visionllmv2} benefits from a specialist detector for spatial localization, yet its outputs often fail to follow the referring expression closely because the detector and LLM operate separately rather than in an end-to-end fashion. In this case, the model failed to identify the ``children to the left".
In contrast, \method is a unified end-to-end model that jointly performs detection and temporal association, resulting in more stable and reliable tracking.

\begin{figure}
    \centering
    \includegraphics[width=0.93\linewidth]{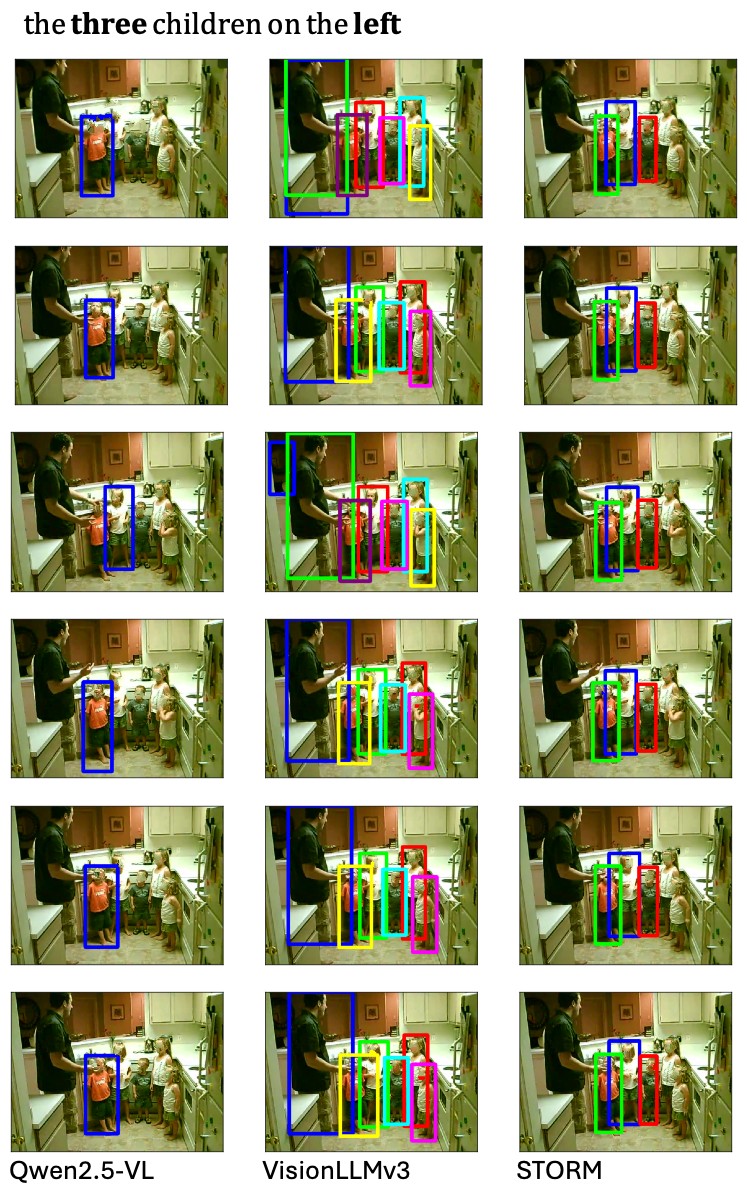}
    \caption{Comparison of \method, Qwen2.5-VL, and VisionLLMv2 on referring multi-object tracking using examples from \dataset. \method\ closely follows the query prompt and accurately localizes and tracks all objects that satisfy the referring expression over time.}
    \label{fig:vis1}
    \vspace{-4mm}
\end{figure}

\vspace{-2mm}
\section{Conclusion}

We present \method, an end-to-end multimodal large language model for referring multi-object tracking that unifies grounding and tracking without external detectors or trackers. Together with task-composition learning and the new \dataset dataset, \method\ achieves state-of-the-art performance across image grounding, single-object tracking, and RMOT benchmarks, demonstrating strong spatial-temporal grounding in challenging real-world videos.

\newpage
{
    \small
    \bibliographystyle{ieeenat_fullname}
    \bibliography{main}
}

\clearpage
\maketitlesupplementary

In the supplementary material, we provide additional visualizations of RMOT model performance and detailed statistics of our \dataset dataset.

\section{Qualitative Results}
\begin{figure*}
    \centering
    \includegraphics[width=0.9\linewidth]{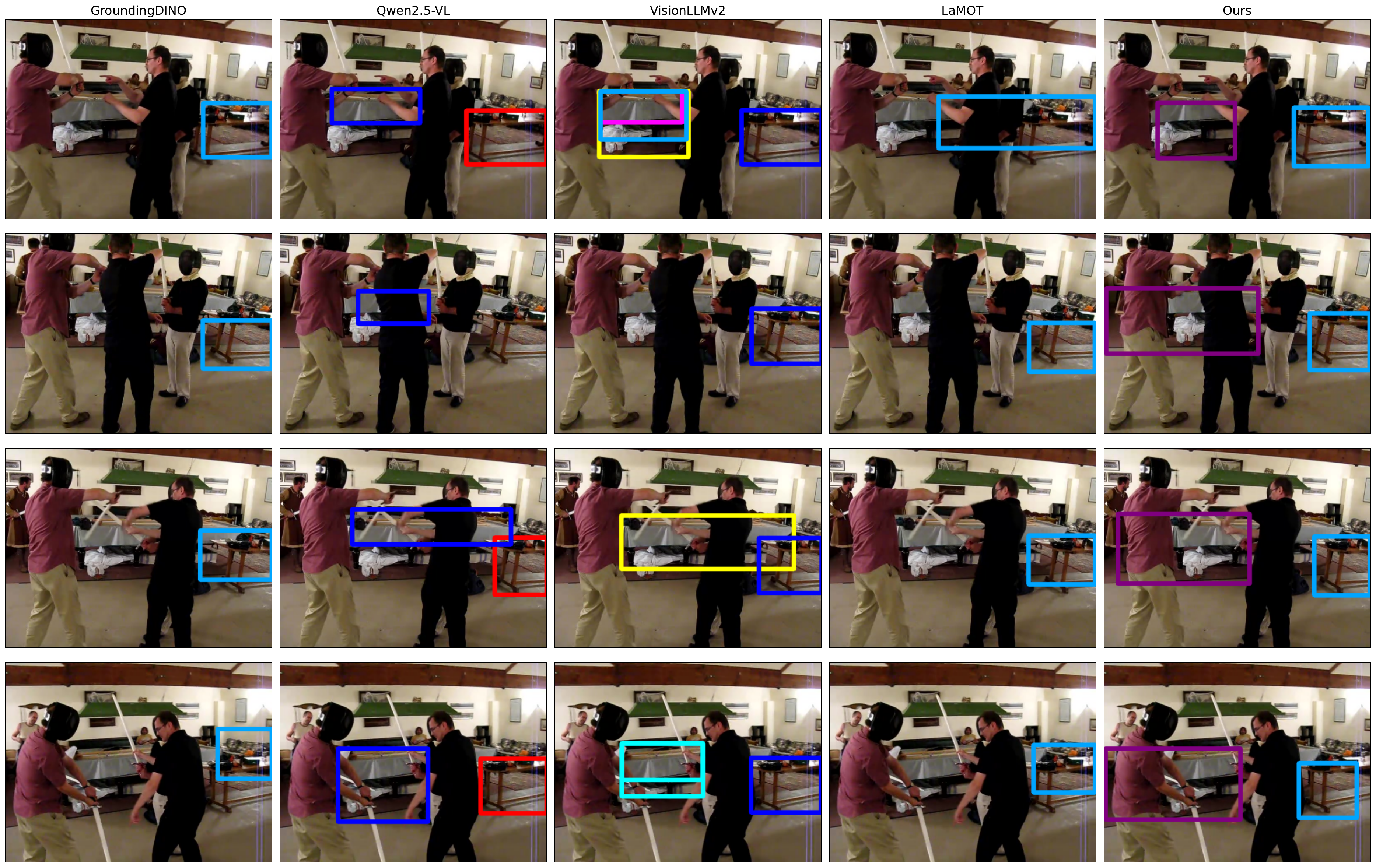}
    \caption{Model results with the referring expression \textbf{the table with white cloth and the wooden table}.}
    \label{fig:vis5}
\end{figure*}

In this section, we present more qualitative examples comparing the referring multi-object tracking performance of our \method with baseline models \cite{li2022grounded,wang2024qwen2,wu2024visionllmv2,li2025lamot} in Figures \ref{fig:vis1}--\ref{fig:vis5}.

Figure \ref{fig:vis1} shows a setting where the model must track both a salient object (the child) and nearby small objects (the toys). Baseline models such as Qwen2.5-VL and LaMOT often miss small objects, while VisionLLMv2 over-detects them, producing imprecise grounding. These issues stem from the fact that most baseline models perform image-based grounding; thus, an object detected in one frame may be missed in subsequent frames, as seen in Qwen2.5-VL. In contrast, \method provides stable predictions and consistently localizes both objects across all frames.

\begin{figure*}
    \centering
    \includegraphics[width=0.9\linewidth]{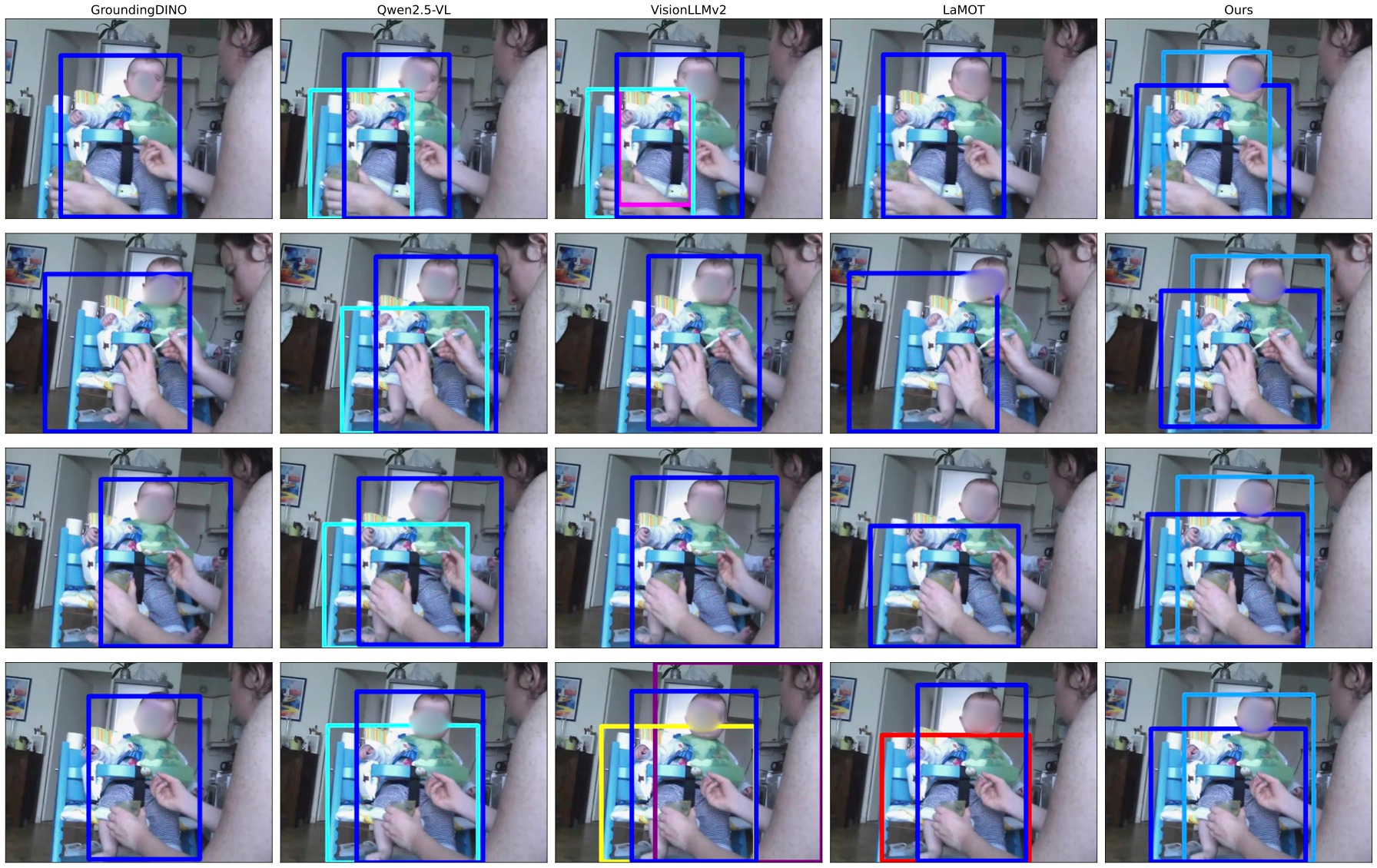}
    \caption{Model results with the referring expression \textbf{the baby and the blue high chair beneath the baby}.}
    \label{fig:vis2}
\end{figure*}

Figure \ref{fig:vis2} illustrates a case with heavy occlusion. The high chair beneath the baby is only partially visible throughout the video. Baselines either miss the chair entirely or ground only a visible sub-part. Occlusion is a common challenge for object detection and tracking, but our task-composition learning strategy allows \method to learn robust occlusion handling from diverse image grounding and single-object tracking data. As a result, \method predicts complete bounding boxes despite partial visibility.

Figure \ref{fig:vis3} shows an example where objects exit and re-enter the frame. Baselines incorrectly produce false-positive bounding boxes in the second frame, even though both the cup and the meat are absent. During training, we enforce that \method outputs an empty bounding box when an object is not visible. Thus, \method does not assume referred objects always appear in the video and can reliably detect their presence or absence.

\begin{figure*}
    \centering
    \includegraphics[width=0.9\linewidth]{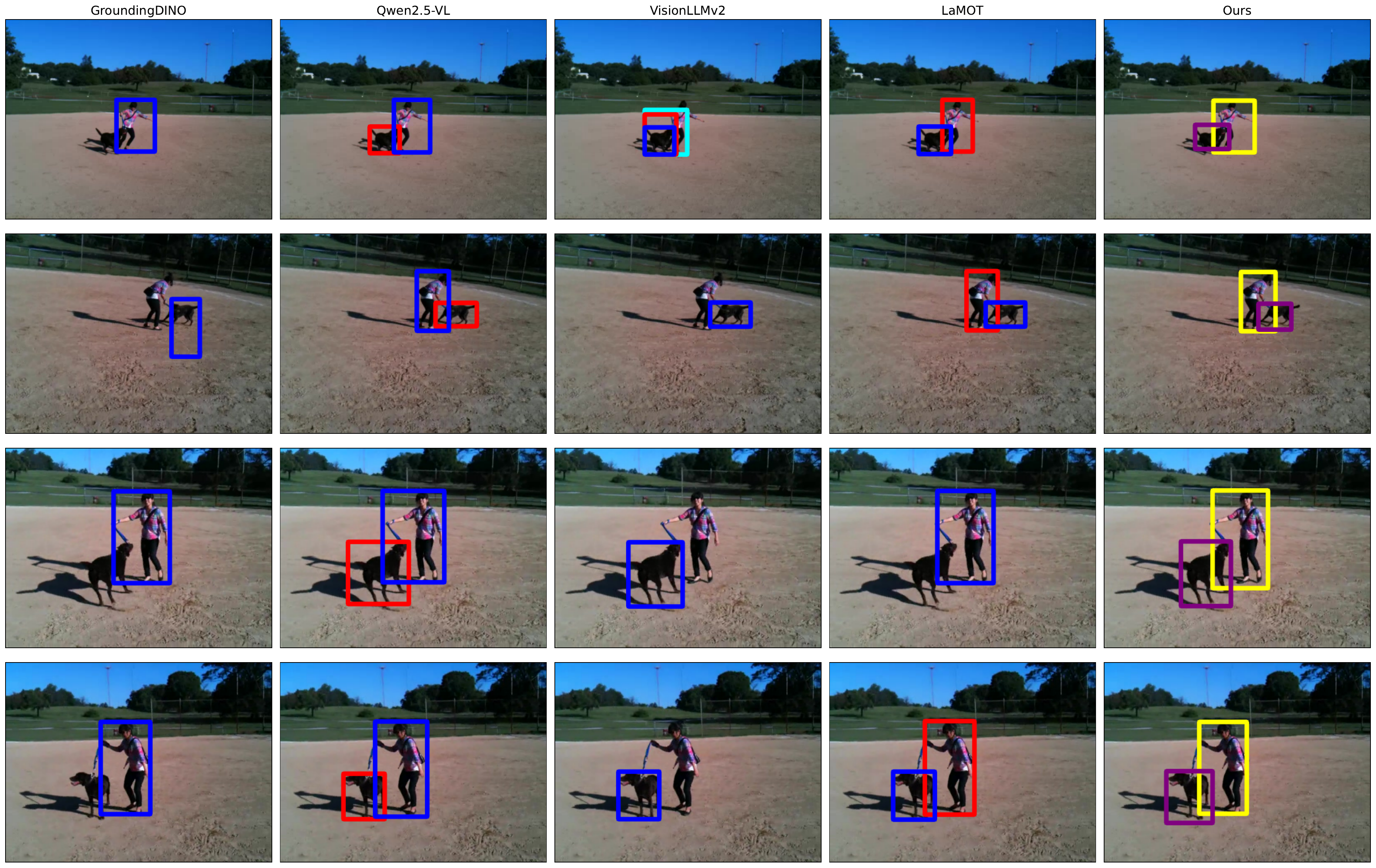}
    \caption{Model results with the referring expression \textbf{the adult wearing a pink and white patterned shirt and the black dog}.}
    \label{fig:vis4}
\end{figure*}

Figure \ref{fig:vis4} presents a scenario with rapid changes in object location and scale. These variations challenge the model to maintain accurate bounding box predictions. GroundingDINO exhibits identity switches between the lady and the dog, whereas \method tracks both objects with consistent identities and stable trajectories.

Finally, Figure \ref{fig:vis5} shows a failure case involving a partially visible table with a white cloth. The crowded scene significantly increases difficulty, leading all models to mislocalize the table. \method misses the right portion of the table. This example highlights that heavy occlusion in dense scenes remains challenging for referring object trackers. We plan to address this limitation in future work by improving training strategies and refining data curation.

\section{Statistics of the \dataset Dataset}
\begin{figure}
    \centering
    \includegraphics[width=\linewidth]{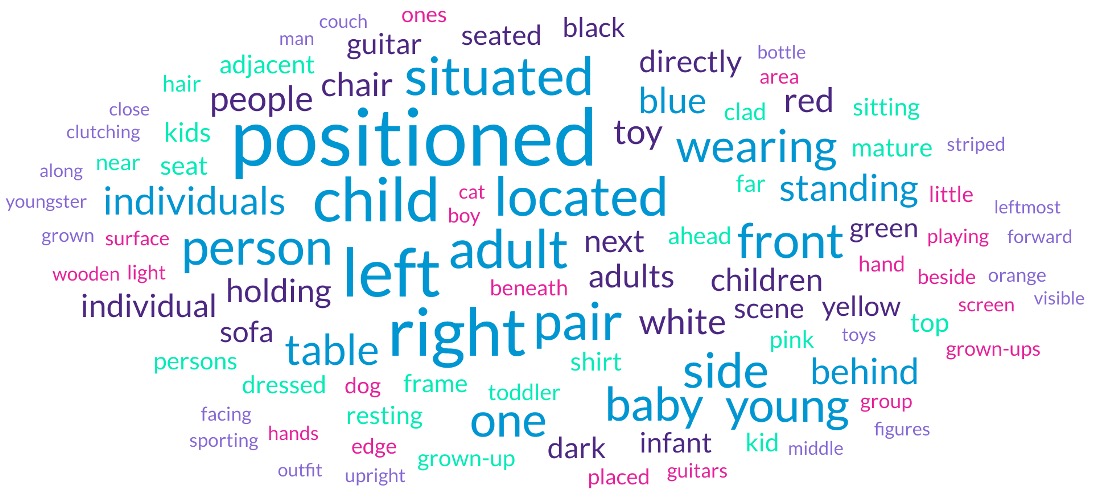}
    \caption{Word cloud of the referring expressions in the \dataset dataset.}
    \label{fig:word}
\end{figure}

\begin{figure*}
    \centering
    \includegraphics[width=\linewidth]{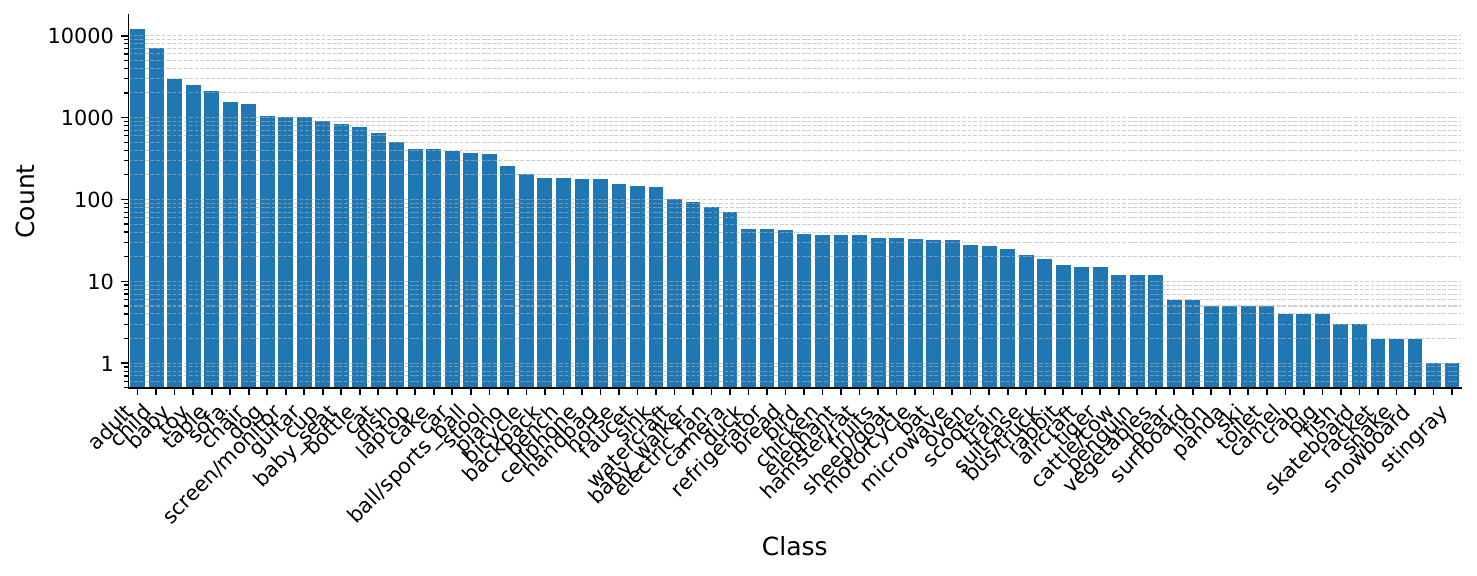}
    \caption{Distribution of object classes in \dataset.}
    \label{fig:dis}
\end{figure*}

In Section 4 of the main paper, we compared the statistics of our dataset with existing referring object tracking datasets. Here, we provide additional visualizations. Figure \ref{fig:word} presents a word cloud of referring expressions in our dataset, showing that expressions commonly describe positional relations (e.g., ``positioned'', ``left''), colors (e.g., ``yellow'', ``pink''), age (e.g., ``adult'', ``toddler''), appearance (e.g., ``outfit'', ``dressed''), and temporal actions or movement (e.g., ``resting'', ``holding''). Figure \ref{fig:dis} visualizes the distribution of object categories, demonstrating that \dataset covers a wide range of everyday objects, including people, animals, furniture, electronics, and food. This stands in contrast to prior MOT datasets \cite{fan2019lasot,milan2016mot16,leal2015motchallenge} that primarily focus on pedestrians and vehicles. Together, these characteristics show that \dataset provides diverse expressions and object types, enabling models to learn richer spatial--temporal grounding and improving generalization across varied real-world scenarios.

\section{Efficiency Analysis}
\change{We further measure inference cost as a function of video length and target count. With a fixed number of targets, decoding remains stable at around 3 FPS as the video grows from 32 to 256 frames. As the number of tracked targets increases, throughput drops roughly linearly because the model must auto-regressively emit longer output sequences.}

\section{Failure Analysis and Limitations}
\change{The high AssA and low IDsw in Table~4 of the main paper indicate that \method\ is already strong at maintaining cross-frame identity consistency. The remaining gap is primarily in grounding quality, as reflected by DetA: the model can still miss precise spatial alignment when prompts are highly ambiguous or when scenes are heavily crowded. We also observe that very long videos or videos with more than ten target objects become challenging because the autoregressive output sequence grows quickly.}

\begin{table}[]
    \centering
    \resizebox{0.8\linewidth}{!}{%
    \begin{tabular}{c|cccc|cccc}
    \toprule
    \#frames & 32 & 64 & 128 & 256 & 64 & 64 & 64 & 64 \\ \midrule
    \#targets & 1 & 1 & 1 & 1 & 1 & 2 & 4 & 8 \\ \midrule
    time (s) & 9.7 & 19.7 & 36.4 & 76.4 & 19.7 & 34.9 & 67.4 & 143.3 \\
    FPS & 3.3 & 3.2 & 3.3 & 3.3 & 3.2 & 1.8 & 1.0 & 0.4 \\
    \bottomrule
    \end{tabular}}
    \caption{\small Efficiency analysis of \method with an 8B MLLM backbone. The first four columns vary video length with one target, while the last four vary the number of targets at 64 frames.}
    \label{tab:efficiency}
    \vspace{-4mm}
\end{table}

\begin{figure*}
    \centering
    \includegraphics[width=0.9\linewidth]{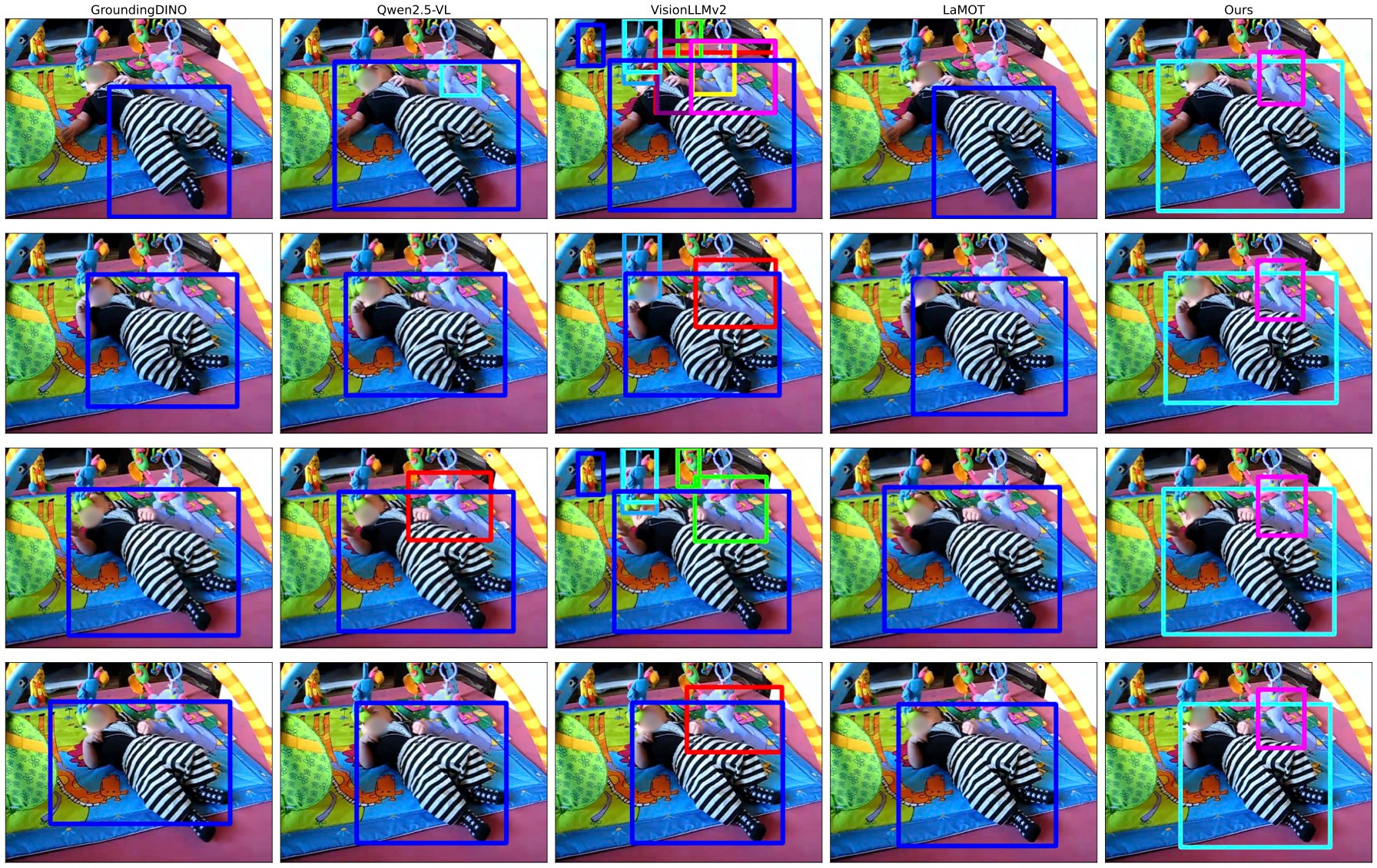}
    \caption{Model results with the referring expression \textbf{the baby and the pink toy above it}.}
    \label{fig:vis1}
\end{figure*}

\begin{figure*}
    \centering
    \includegraphics[width=0.9\linewidth]{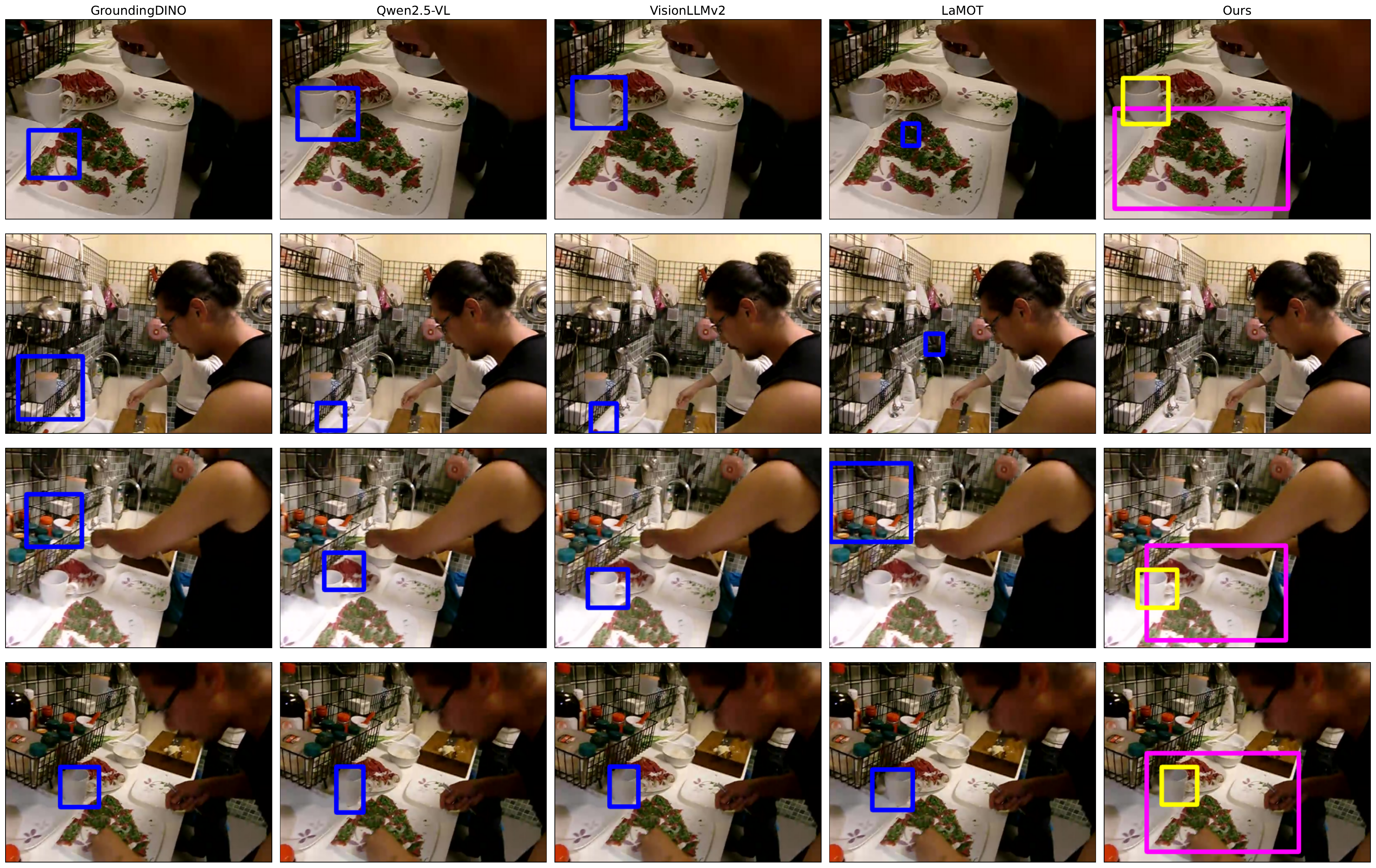}
    \caption{Model results with the referring expression \textbf{the cup next to the dish with fan-shaped meat arrangement}.}
    \label{fig:vis3}
\end{figure*}

\end{document}